%% file: paper.tex
\documentclass[]{spie}  %>>> use for US letter paper
\usepackage{amsmath,amsfonts,amssymb, bbm}
\usepackage{graphicx,subfig,float}
\usepackage{booktabs}
\usepackage{xcolor}
\usepackage{colortbl}
\usepackage{multirow}
\usepackage{hhline}
\usepackage{caption}
\usepackage{nicefrac}
\usepackage[]{algorithm2e}
\usepackage[colorlinks=true, allcolors=blue]{hyperref}

\title{Self-Training with Improved Regularization for Sample-Efficient Chest X-Ray Classification}

\author[a]{Deepta Rajan}
\author[b]{Jayaraman J. Thiagarajan}
\author[a]{Alexandros Karargyris}
\author[a]{Satyananda Kashyap}
\affil[a]{IBM Research Almaden}
\affil[b]{Lawrence Livermore National Labs}

\authorinfo{Further author information: (Send correspondence to Deepta Rajan)\\Deepta Rajan: E-mail: r.deepta@gmail.com; Jayaraman J. Thiagarajan: E-mail: jjthiagarajan@gmail.com; Alexandros Karargyris: E-mail: akarargyris@gmail.com; Satyananda Kashyap: E-mail: Satyananda.Kashyap@ibm.com}

\begin{document} 
\maketitle

\begin{abstract}
Automated diagnostic assistants in healthcare necessitate accurate AI models that can be trained with limited labeled data, can cope with severe class imbalances and can support simultaneous prediction of multiple disease conditions. To this end, we present a deep learning framework that utilizes a number of key components to enable robust modeling in such challenging scenarios. Using an important use-case in chest X-ray classification, we provide several key insights on the effective use of data augmentation, self-training via distillation and confidence tempering for small data learning in medical imaging. Our results show that using $85\%$ lesser labeled data, we can build predictive models that match the performance of classifiers trained in a large-scale data setting.  
\end{abstract}


\section{INTRODUCTION}
\label{sec:intro}  % \label{} allows reference to this section
\input{sections/intro}

\section{PROBLEM SETUP}
\label{sec:problem}
\input{sections/problem}

\section{METHODS}
\label{sec:method}
\input{sections/method}

\section{RESULTS \& FINDINGS}
\label{sec:results}
\input{sections/results}

\section{DISCUSSION}
\label{sec:discussion}
\input{sections/discussion}

\section{ACKNOWLEDGEMENTS}
This work was performed under the auspices of the U.S. Department of Energy by Lawrence Livermore National Laboratory under Contract DE-AC52-07NA27344.

% References
\bibliographystyle{spiebib}
\bibliography{refs} % bibliography data in report.bib

\end{document}

%% file: sections/intro.tex
An increasing need for automated diagnostic assistants in healthcare places a growing demand for developing accurate AI models, while being resilient to biases stemming from data sources and demographics~\cite{thiagarajan2018can}. In this paper, we consider an important class of diagnosis problems in medical imaging that is characterized by three crucial real-world challenges: (i) limited access to labeled data, (ii) severe class imbalance, and (iii) the need to associate each sample to multiple disease conditions (multi-label). Learning with limited labeled data, when combined with imbalanced sampling, leads to severe overfitting in practice. Though the recent advances to few-shot learning can help with this challenge to an extent, e.g. novel augmentation techniques~\cite{chaitanya2019semi,eaton2018improving}, customized loss functions~\cite{ge2018chest} and sophisticated regularization strategies~\cite{zhang2017mixup}, the class imbalance and multi-label nature of clinical diagnosis problems make them insufficient in practice. Another popular approach to deal with small data problems is to leverage additional unlabeled datasets, if available, and build more robust models~\cite{berthelot2019mixmatch,berthelot2019remixmatch,arazo2019pseudo}. However, their effectiveness in several clinical benchmark problems has not been well-studied so far. 

\noindent \textbf{Use-case.} To illustrate the aforementioned challenges, we consider a chest X-ray (CXR) classification problem, and show that existing deep learning-based solutions developed for large-scale data perform poorly with limited data~\cite{rajpurkar2018deep,pham2019interpreting}. More specifically, we use the public CXR repository developed by Stanford~\cite{irvin2019chexpert}. The choice of this use-case was driven by the prevalence of X-rays as a diagnostic modality~\cite{mettler2009radiologic}, the impact of robustly detecting lung conditions~\cite{xu2020pathological} and the difficulty in obtaining expert annotations at scale~\cite{irvin2019chexpert}.

\noindent \textbf{Proposed Work.} In this paper, we develop a deep learning framework for training robust medical image classification models in very low sample regimes. More specifically, our framework combines the following learning strategies to improve prediction performance in scenarios characterized by limited labeled data availability and severe class imbalance: (i) \textit{weak image augmentation}: pre-defined image transformations such as rotation, shift etc. to synthesize additional examples; (ii) \textit{mixup training}: regularization based on Vicinal Risk Minimization~\cite{zhang2017mixup}; (iii) a novel \textit{confidence tempering} regularization to handle the class imbalance challenge in multi-label classification problems; and (iv) \textit{distillation-based self-training} with noisy students. While image augmentation is routinely used in state-of-the-art solutions for CXR classification~\cite{bressem2020comparing}, we make a surprising finding that, with small data, it is insufficient to achieve good generalization just on its own. Hence, we propose to adopt both mixup training and confidence tempering to improve model generalization. In our experiments, we find that these regularization strategies together with weak augmentation lead to significant performance gains. 

An inherent challenge in limited data problems is that the resulting models contain non-generalizable inductive biases. Hence, we introduce a distillation-based self-training protocol that can leverage additional unlabeled data (when available) to systematically adjust the biases in the \textit{teacher} model (predictive model trained using the aforementioned regularization strategies) and evolve a \textit{student} model that can generalize better than the teacher. We make a crucial finding, similar to~\cite{xie2019self}, that a \textit{noisy} student leads to better performance. Though our approach uses prediction sharpening and mixup training to create noisy students, other approaches including dropout or stochastic depth can also be used. 

\noindent \textbf{Findings.}
Our results show that a ResNet-18 model trained using our proposed approach with only $12.5$k labeled ($<10\%$ of the dataset) and $15$k unlabeled samples outperforms a state-of-the-art ResNet-18 model trained on the complete $138$k labeled set. More interestingly, when the amount of available labeled data is increased to $\sim15\%$, our approach achieves comparable performance to more sophisticated architectures including ResNet-50 and DenseNet-121 trained on the entire dataset. 

%% file: sections/problem.tex
In this work, we consider the problem of chest X-ray classification to find evidences for any combination of $5$ different diseases, namely: (a) Atelectasis (AT), (b) Cardiomegaly (CA), (c) Consolidation (CO), (d) Edema (ED), and (e) Pleural Effusion (PE). In our setup, we assume that we can only access limited labeled data and the label distribution is characterized by severe imbalance.
\begin{table}[t]
	\centering
	\renewcommand{\arraystretch}{1.3}
	\renewcommand{\tabcolsep}{5pt}
	\begin{tabular}{l|c|c|c|c|c|c|c}
		\hline
		\multicolumn{1}{c|}{\cellcolor{gray!15}\textbf{Dataset}} &
		\cellcolor{gray!15}\textbf{Patients} & \cellcolor{gray!15}\textbf{Images} & \cellcolor{gray!15}\textbf{CA} &
		\cellcolor{gray!15}\textbf{ED} &
		\cellcolor{gray!15}\textbf{CO} &
		\cellcolor{gray!15}\textbf{AT} &
		\cellcolor{gray!15}\textbf{PE} \\ \hline
		Train  & 43,393 & 138,655  & 17,572 & 36,983 & 10,040 & 23,810 & 58,141  \\ 
		Validation & 10,000 & 20,674 & 1,849 & 3,543 & 1,016 & 3,337 & 5,632  \\  
		Test & 200 & 234 & 68 & 45 & 33 & 80 & 67   \\ \hline
	\end{tabular}
	\caption{Description of the chest X-ray classification dataset used in our study.}
	\label{table:data-size}
\end{table}

\noindent \textbf{Dataset Description.} We use CheXpert~\cite{irvin2019chexpert}, a large public dataset for chest radiograph interpretation. The images were curated by Stanford from both in-patient and out-patient centers between October 2002 and July 2017. It consists of $224,316$ X-rays from $65,240$ patients, where images can correspond to Frontal, Lateral, Anteroposterior or Posteroanterior views. In our study, we used the subset of the train set that contained an actual prediction for the $5$ classes that we considered (some of the samples have the label \textit{uncertain}). Subsequently, we randomly split the dataset into train and validation sets with no patient overlap among them and the test set was designed using the additional $200$-patient set released publicly by Stanford for evaluation. The sample sizes used in our experiments along with their class distributions are summarized in Table~\ref{table:data-size}.

\noindent \textbf{Setup.}
We denote a labeled dataset by the tuple, $(\mathrm{X}_{\ell},\mathrm{Y}_{\ell})$, which is a collection of $N_{\ell}$ examples and a label matrix of size $N_{\ell} \times \mathrm{C}$, where $\mathrm{C}$ indicates the total number of classes (set to $5$). We denote the additional unlabeled dataset of $N_u$ samples as $\mathrm{X}_u$, and we do not have access to their corresponding annotations. In our experiments, we randomly draw both labeled and unlabeled sets from the $138$k train set (see Table~\ref{table:data-size}) with no overlap between them. Note, we assume that the marginal distributions of the $5$ classes in the labeled dataset $(\mathrm{X}_{\ell},\mathrm{Y}_{\ell})$ is same as the original $138$k training set. We expect the classification task to be significantly more challenging as the number of labeled examples $N_{\ell}$ becomes smaller. 

In order to use models pre-trained on ImageNet for initialization, we pre-processed the raw gray-scale images by resizing them to $224 \times 224 \times 3$ using linear interpolation while maintaining the aspect ratio with border padding. The images were then normalized using a pixel mean of $128.0$ and standard deviation of $64.0$ in addition to being contrast adjusted using histogram equalization.

%% file: sections/method.tex
\begin{figure*}[t]
	\centering
	\includegraphics[width=0.9\linewidth]{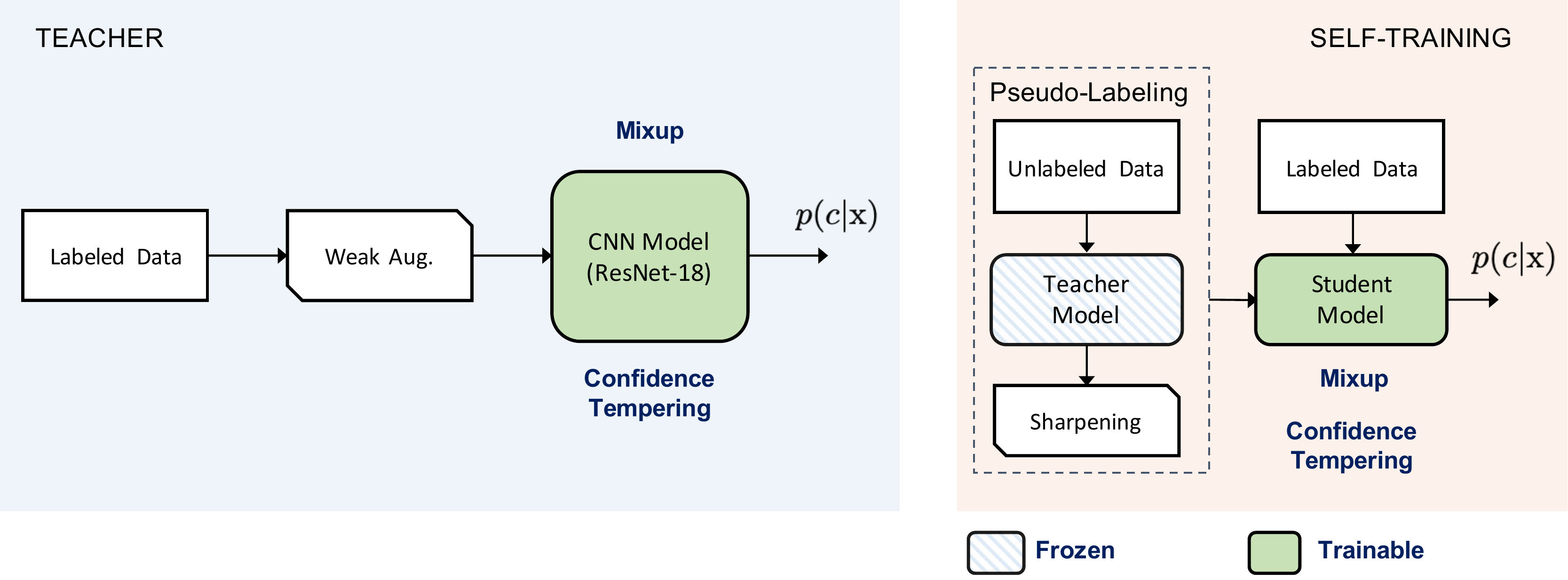}
	\vspace{0.1in}
	\caption{An overview of the proposed small data learning framework.}
	\label{fig:overview}
\end{figure*}

In this section, we describe the proposed approach for building accurate classifiers using limited labeled data. As discussed in Section \ref{sec:intro}, our approach is comprised of four crucial components, namely (i) weak image augmentation, (ii) mixup training, (iii) confidence tempering, and (iv) self-training with noisy students. Figure \ref{fig:overview} illustrates an overview of the proposed approach.

\noindent \textbf{(i) Weak image augmentation.} In accordance with one of the CheXpert~\cite{irvin2019chexpert} competition's top-ranked submission~\cite{jfhealthcare}, we perform weak augmentation on X-rays to improve robustness of the trained models. In particular, we apply random affine transformations namely rotation ($-15^{\circ}$ to $15^{\circ}$), horizontal/ vertical translations ($-0.05$ to $0.05$) and scaling ($0.95$ to $1.05$). Though weak data augmentation is widely adopted to avoid overfitting, we find it to be insufficient in problems with limited training data, corroborating with the results in~\cite{chaitanya2019semi}, where other augmentation techniques were also explored. However, from our ablation studies, we find that when coupled with additional regularization strategies, weak augmentation can be useful even for limited data settings. In the rest of the paper, we refer to the augmented images by the notation $\bar{\mathrm{X}}_{\ell}$.  

\noindent \textbf{(ii) Mixup training.}
Mixup is a popular technique for training deep neural networks~\cite{zhang2017mixup}, wherein we generate additional samples by convexly combining random pairs of input images and their corresponding labels. It is based on the principle of Vicinal Risk Minimization~\cite{chapelle2001vicinal}, where the goal is to train classifiers not only on the training samples, but also in the vicinity of each sample. It has also been found in~\cite{thulasidasan2019mixup} that mixup training leads to networks whose confidences are well-calibrated, i.e., the predictive scores are actually indicative of the actual likelihood of correctness. To this end, we create virtual image-label pairs by convexly interpolating between two random samples $\{(\bar{\mathrm{x}}_i, \mathrm{y}_i), (\bar{\mathrm{x}}_j, \mathrm{y}_j)\}$, where $\bar{\mathrm{x}}_i, \bar{\mathrm{x}}_j \in \bar{\mathrm{X}}_{\ell}$,
\begin{align}
    \mathrm{\tilde{x}} = \lambda\bar{\mathrm{x}}_i+(1-\lambda)\bar{\mathrm{x}}_j; 
    \mathrm{\tilde{y}} = \lambda\mathrm{y}_i+(1-\lambda)\mathrm{y}_j
    \label{eqn:mixup}
\end{align}and enforce the consistency that predictions for $\mathrm{\tilde{x}}$ should agree with the interpolated labels $\tilde{\mathrm{y}}$. The amount of interpolation is controlled by the parameter $\alpha$, where $\alpha \in (0,\infty)$ in $\lambda \sim Beta(\alpha,\alpha)$, and $Beta$ denotes the beta distribution. In practice, given the predictions from a model $\mathcal{F}$ with parameters $\mathrm{\theta}$, we define the loss function for mixup training as follows:
\begin{equation}
\mathcal{L}_{mixup}(\bar{\mathrm{X}}_{\ell},\mathrm{Y}_{\ell};\mathcal{F}) = \sum_{\{(\bar{\mathrm{x}}_i, \mathrm{y}_i), (\bar{\mathrm{x}}_j, \mathrm{y}_j)\} \in \bar{\mathrm{X}}_{\ell},\mathrm{Y}_{\ell}} \lambda \mathcal{L}_{bce}(\tilde{\mathrm{x}},\mathrm{y}_i; \mathcal{F} ) + (1-\lambda) \mathcal{L}_{bce}(\tilde{\mathrm{x}},\mathrm{y}_j; \mathcal{F} ),
    \label{eqn:loss_m}
\end{equation}where $\mathcal{L}_{bce}(\tilde{\mathrm{x}},\mathrm{y}_i; \mathcal{F} )$ denotes the binary cross entropy loss between the predictions $\mathcal{F}(\tilde{\mathrm{x}})$ and the true labels $\mathrm{y}_i$, and the summation is over multiple random pairs.

\RestyleAlgo{boxruled}
\begin{algorithm}[t]
	\KwIn{Labeled data $(\mathrm{X}_{\ell}, \mathrm{Y}_{\ell})$, and unlabeled data $(\mathrm{X}_u)$, Mixup parameter $\alpha$, confidence tempering constants $\tau_l$ and $\tau_h$, sharpening parameter $\gamma$, hyper-parameters $\beta_c, \beta_e^{\ell}, \beta_e^u, \beta_c^{u}$.}
	\KwOut{Teacher model $\mathcal{F}$ with parameters $\mathrm{\theta}^*$ and Student model $\mathcal{G}$ with parameters $\mathrm{\phi}^*$}
	\textbf{Initialization}: initialize model parameters $\mathrm{\theta}$\;
	
	    \For{$n$ epochs}{
        	    Perform weak image augmentation to labeled data to obtain $\bar{\mathrm{X}}_{\ell}$\;
        	    Generate mixup parameter $\lambda \sim Beta(\alpha,\alpha)$\;
                \textit{Mixup training}: Convexly combine random sample pairs in $\bar{\mathrm{X}}_{\ell}$ using eqn. \eqref{eqn:mixup} and compute $\mathcal{L}_{mixup}(\bar{\mathrm{X}}_{\ell},\mathrm{Y}_{\ell})$ using eqn. \eqref{eqn:loss_m}\;
                \textit{Confidence tempering}: For each class $\mathrm{\mathrm{c}}$, estimate $\mathcal{R}_{ct}(c)$ using eqn. \eqref{eqn:ct}\; 
                Compute loss function $\mathcal{L} = \mathcal{L}_{mixup}(\bar{\mathrm{X}}_{\ell},\mathrm{Y}_{\ell}) + \beta_c \sum_{\mathrm{c}} \mathcal{R}_{ct}(\mathrm{c})$\;
                Update parameters $\mathrm{\theta}^* = \arg \min_{\mathrm{\theta}} \mathcal{L}$\;
            }
            /*\textit{Self-training}*/ \\
            Initialize a student model $\mathcal{G}$ with parameters $\mathrm{\phi}$\;
            \For {m \textit{epochs}}{
            Perform weak image augmentation to labeled and unlabeled data to obtain $\bar{\mathrm{X}}_{\ell}$ and $\bar{\mathrm{X}}_u$\;
            Estimate pseudo labels for unlabeled data $\hat{\mathrm{Y}}_u = \mathcal{F}(\bar{\mathrm{X}}_u; \mathrm{\theta}^*)$\;
            Perform sharpening of $\hat{\mathrm{Y}}_u$ with $\gamma$ using eqn. \eqref{eqn:sharp}\;
            Generate mixup parameter $\lambda \sim Beta(\alpha,\alpha)$\;
            \textit{Mixup training for distillation}: Convexly combine random pairs in $\bar{\mathrm{X}}_u$ using eqn. \eqref{eqn:mixup} and compute $\mathcal{L}_{dist}(\bar{\mathrm{X}}_u,\hat{\mathrm{Y}}_{u,\gamma})$ using eqn. \eqref{eqn:loss_dist}\;
            \textit{Confidence tempering}: For each $\mathrm{\mathrm{c}}$, estimate $\mathcal{R}^{u}_{ct}(c)$ using eqn. \eqref{eqn:ct} for $\bar{\mathrm{X}}_u$\; 
            Compute loss function $\mathcal{L}_s = \beta_e^{\ell}\mathcal{L}_{bce}(\bar{\mathrm{X}}_{\ell},\mathrm{Y}_{\ell}) + \beta_e^{u} \mathcal{L}_{dist}(\bar{\mathrm{X}}_u,\hat{\mathrm{Y}}_{u,\gamma}) + \beta_c^{u} \sum_{\mathrm{c}} \mathcal{R}_{ct}^u(\mathrm{c})$\;
            Update parameters $\mathrm{\phi}^* = \arg \min_{\mathrm{\phi}} \mathcal{L}_s$\;}
        \textbf{return} $\mathcal{F}(\mathrm{\theta}^*), \mathcal{G}(\mathrm{\phi}^*)$\;
        
	\caption{Proposed approach with limited labeled data and an additional unlabeled set.}\label{algo-d}
\end{algorithm}

\noindent \textbf{(ii) Confidence tempering regularization.} Though mixup training helps in avoiding model overfitting, the inherent class imbalance can make it ineffective, particularly for classes with lesser number of examples. A na\"ive way to handle this is to alter the probability distribution with which we choose the random pairs in mixup (i.e. uniform distribution), however, it is not clear how to estimate marginal distributions using limited data that effectively reflects the unseen test cases. Hence, we propose a novel regularization strategy, referred as \textit{confidence tempering (CT)}. A common observation in imbalanced multi-label problems is that a model compounds more evidence for assigning every image to the most prevalent class, while providing little to no likelihood for classes with very few examples. We avoid this by introducing a regularization term for every class $c$:
\begin{equation}
\mathcal{R}_{ct}(\mathrm{c}) = \log\left(\frac{\tau_{l}}{\rho_{\mathrm{c}}} + \frac{\rho_{\mathrm{c}}}{\tau_{h}}\right), \text{where } \rho_{\mathrm{c}} = \frac{1}{N} \sum_{i=1}^N p_i(\mathrm{c}).
\label{eqn:ct}
\end{equation}Here, $p_i(\mathrm{c})$ indicates the likelihood of assigning sample $\mathrm{x}_i$ to class $\mathrm{c}$ and $\rho_{\mathrm{c}}$ is the average evidence for class $\mathrm{c}$. In practice, we evaluate this for each mini-batch. In other words, this regularization penalizes a model that assigns overwhelmingly high evidences for any class or that does not provide any non-trivial evidence for any class. As we will show in our results, this regularization provides significant performance gains for classes with very few examples in the train set. The hyper-parameters $\tau_l$ and $\tau_h$ are low and high thresholds for tempering confidences and were set to $0.35$ and $0.75$ respectively (based on a hyper-parameter search). 

\noindent \textbf{(iv) Self-training with noisy students.} 
%\subsection{Self-training with Noisy Students}
Finally, we propose to employ a self-training protocol, wherein we distill knowledge from a trained model $\mathcal{F}(\mathrm{\theta})$ (\textit{Teacher}) to evolve a \textit{Student} model $\mathcal{G}$ with parameters $\mathrm{\phi}$ that can achieve an improved generalization. We assume that we have access to unlabeled data $\mathrm{X}_u$, in addition to the limited labeled set. This assumption well reflects the real-world scenarios, wherein obtaining high-quality annotations is the more expensive and time-consuming step. As illustrated in Figure \ref{fig:overview}, the teacher model $\mathcal{F}$ is trained solely using the labeled data (along with weak augmentation, mixup and confidence tempering). Subsequently, the unlabeled data is used to adjust the inductive biases and evolve the next generation model $\mathcal{G}$ (student) that has better generalization characteristics. We build upon the empirical evidence in the recent work by Xie \textit{et al.}~\cite{xie2019self} and use a student model that is noised during training. While the authors in~\cite{xie2019self} used dropout to noise the student model, we use mixup regularization to implement a noisy student.

Formally, given the teacher model $\mathcal{F}(\mathrm{\theta}^*)$, we first estimate pseudo-labels for the weakly augmented unlabeled data, $\hat{\mathrm{Y}}_u = \mathcal{F}(\bar{\mathrm{X}}_u; \mathrm{\theta}^*)$. In order to reduce the effect of uncertainties in the teacher model, we perform sharpening of the predictions as follows:
\begin{equation}
    \hat{\mathrm{Y}}_{u,\gamma} = (1-\gamma)\hat{\mathrm{Y}}_u + \gamma  \mathbbm{1}[\hat{\mathrm{Y}}_u \geq 0.5],
    \label{eqn:sharp}
\end{equation}where $\mathbbm{1}$ denotes the indicator function and $\gamma$ is a hyper-parameter. In practice, to make it differentiable, we implement the indicator function as $\texttt{Sigmoid}(1e8\times(\hat{\mathrm{Y}}_u - 0.5))$. This sharpening pushes the prediction probabilities for each of the labels closer to $1$ when it is greater than $0.5$, and closer to $0$ when it is less than $0.5$. This formulation for multi-label predictions is akin to temperature scaling for multi-class problems, and we set $\gamma = 0.5$ in our experiments. Using the true-labels for the labeled set $\bar{\mathrm{X}}_{\ell}$ and the pseudo labels for the unlabeled set $\bar{\mathrm{X}}_u$, we update the student model parameters $\mathrm{\phi}$. More specifically, we use the following loss function:
\begin{equation}
\mathcal{L}_s = \beta_e^{\ell}\mathcal{L}_{bce}(\bar{\mathrm{X}}_{\ell},\mathrm{Y}_{\ell}) + \beta_e^{u} \mathcal{L}_{dist}(\bar{\mathrm{X}}_u,\hat{\mathrm{Y}}_{u,\gamma}) + \beta_c^{u} \sum_{\mathrm{c}} \mathcal{R}_{ct}^u(\mathrm{c}).
    \label{eqn:student}
\end{equation}The first term is the standard binary cross entropy loss on the labeled set and it ensures that student model is consistent with the known annotations. The second term is the distillation cost that measures the discrepancy between the student predictions for the unlabeled data and the corresponding (sharpened) predictions from the teacher. Similar to existing knowledge distillation approaches~\cite{hinton2015distilling}, we use the KL-divergence to measure the discrepancy.  Since we want to make the student noisy (i.e., introduce uncertainties), we implement the distillation cost as a mixup training objective. More specifically,
\begin{equation}
\mathcal{L}_{dist}(\bar{\mathrm{X}}_u,\hat{\mathrm{Y}}_{u,\gamma}) = \sum_{\{(\bar{\mathrm{x}}_i, \mathrm{y}_i), (\bar{\mathrm{x}}_j, \mathrm{y}_j)\} \in \bar{\mathrm{X}}_u,\hat{\mathrm{Y}}_{u,\gamma}} \lambda KL(\mathrm{y}_i, \mathcal{G}(\tilde{\mathrm{x}})) + (1-\lambda) KL(\mathrm{y}_j, \mathcal{G}(\tilde{\mathrm{x}})),
    \label{eqn:loss_dist}
\end{equation}where $KL$ denotes the KL-divergence. The last term is the confidence tempering regularization of the student predictions for the unlabeled data. We perform confidence tempering only for the unlabeled set during student training. A detailed listing of our approach can be found in Algorithm \ref{algo-d}.

%% file: sections/results.tex
\begin{table}[t]
\centering
\caption{Performance comparison between ResNet-18 models trained using different strategies in the proposed framework: Weak Augmentation (W-Aug.), Mixup, Confidence Tempering (CT) and Self-Training with additional unlabeled data (ST).} 
\renewcommand{\arraystretch}{1.3}
\renewcommand{\tabcolsep}{1.0pt}
\begin{tabular}{|c||c|c||c|c||c|c|}
\hline
\multirow{2}{*}{\textbf{Method}} &
  \multicolumn{2}{c||}{$\mathbf{N_{\ell} = 1000}$} &
  \multicolumn{2}{c||}{$\mathbf{N_{\ell} = 12500}$} &
  \multicolumn{2}{c|}{$\mathbf{N_{\ell} = 20000}$} \\
  \cline{2-7}
                    & W-AUC      & W-PRC     & W-AUC     & W-PRC     & W-AUC     & W-PRC     \\ \hline \hline
W-Aug.              & 0.724      & 0.478     & 0.814     & 0.615     & 0.831          &  0.670         \\
W-Aug. + Mixup      & 0.733      & 0.502     & 0.819     & 0.640      &  0.842         & 0.684           \\
W-Aug. + Mixup + CT & 0.738      & 0.507     & 0.833     & 0.673     & 0.841       & \textbf{0.691}         \\
% W-Aug. + Mixup + CT + ST(L)&
%   0.741 &
%   \textbf{0.542} &
%   0.839 &
%   0.670 & 0.838
%   & 0.688
%   \\
W-Aug. + Mixup + CT + ST &
  \textbf{0.75} &
  \textbf{0.538} &
  \textbf{0.844} &
  \textbf{0.684} & \textbf{0.846}
   & 0.688
   \\ \hline 
%   \hline
%   \rowcolor{gray!10}
% Full Training ($N_{\ell} = 138,655$)       & \multicolumn{3}{c|}{W-AUC = 0. 859} & \multicolumn{3}{c|}{W-PRC = 0.675}\\
% \hline
\end{tabular}
\label{tab:results}
\end{table}

\noindent \textbf{Model Design.} Since the release of CheXpert~\cite{irvin2019chexpert}, a plethora of approaches have been published for X-ray classification including CheXNext~\cite{rajpurkar2018deep}. While most successful solutions use over-parameterized, deep models such as DenseNet-121~\cite{pham2019interpreting}, more recently, even shallow network architectures have been shown to produce comparable performances~\cite{bressem2020comparing}. Given our small data learning setup, we find ResNet-18 to be effective in avoiding overfitting without trading-off performance~\cite{ge2018chest}. We refer to the case where we fine-tune ResNet-18 with only weak augmentation (W-Aug.) as the \textit{baseline} solution. For the proposed approach, we create variants by ablating different components in Algorithm~\ref{algo-d}.

% Note, we compare our proposed approach (with few-shots) to over-parameterized models such as DenseNet-121 trained on the full $138$k training data.

\noindent \textbf{Training.} All models in our study were implemented using Pytorch and trained for $15$ epochs using the following hyperparameters: learning rate $1e-4$ reduced by a factor of $0.1$, batch size $100$, the Adam optimizer with weight decay $1e{-4}$ and momentum $0.9$. For $\alpha$ in Eq.~\eqref{eqn:mixup}, we chose the best values in the range $0.2-0.4$, while a higher $\alpha = 0.6$ works better for $\mathrm{N_\ell}=1000$. We set $\beta_e^{u}$ to $0.8$ in Eq.~\eqref{eqn:student}, and chose the best values between $0.1$ and $0.25$ for $\beta_c$ and $\beta_c^{u}$. Note, we varied $\mathrm{N_{\ell}}=\{1000, 12500, 20000\}$ and the unlabeled set was fixed to be the same for all cases with $\mathrm{N_u}=15000$. The software codes for implementing our framework is available at \href{https://github.com/Drajan/SelfTrain-ChestX-rays}{github.com/Drajan}.

\noindent \textbf{Evaluation Metrics.} To evaluate performance, we use the widely-adopted metrics, namely area under ROC curve (AUC) and precision-recall curve (PRC). Due to the inherent class imbalance, we used weighted averages of the metrics using class-specific weights, which we refer to as W-AUC and W-PRC respectively.

\begin{figure*}[t]
	\centering
	\includegraphics[width=0.95\linewidth]{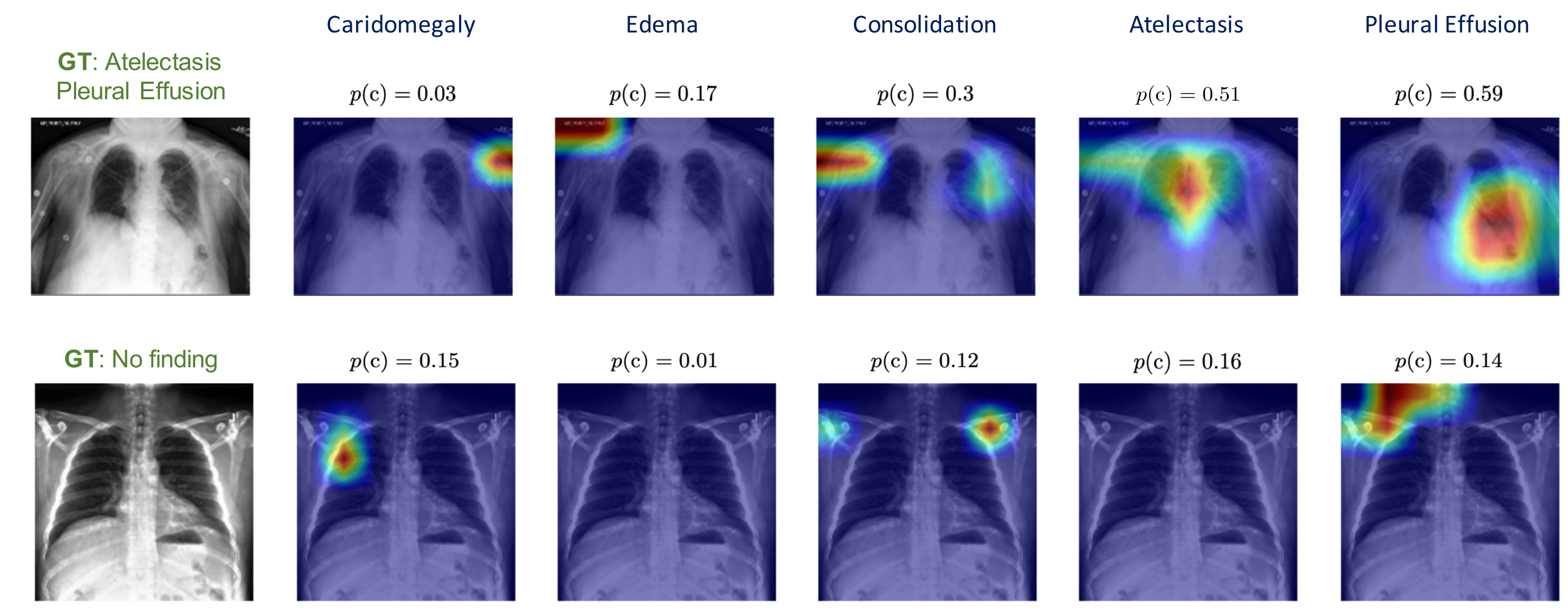}
	\vspace{0.1in}
	\caption{Class-activation maps for two test cases: true positive and true negative.}
	\label{fig:gradcam}
\end{figure*}
\begin{figure*}[t]
	\centering
	{\includegraphics[width=0.8\linewidth]{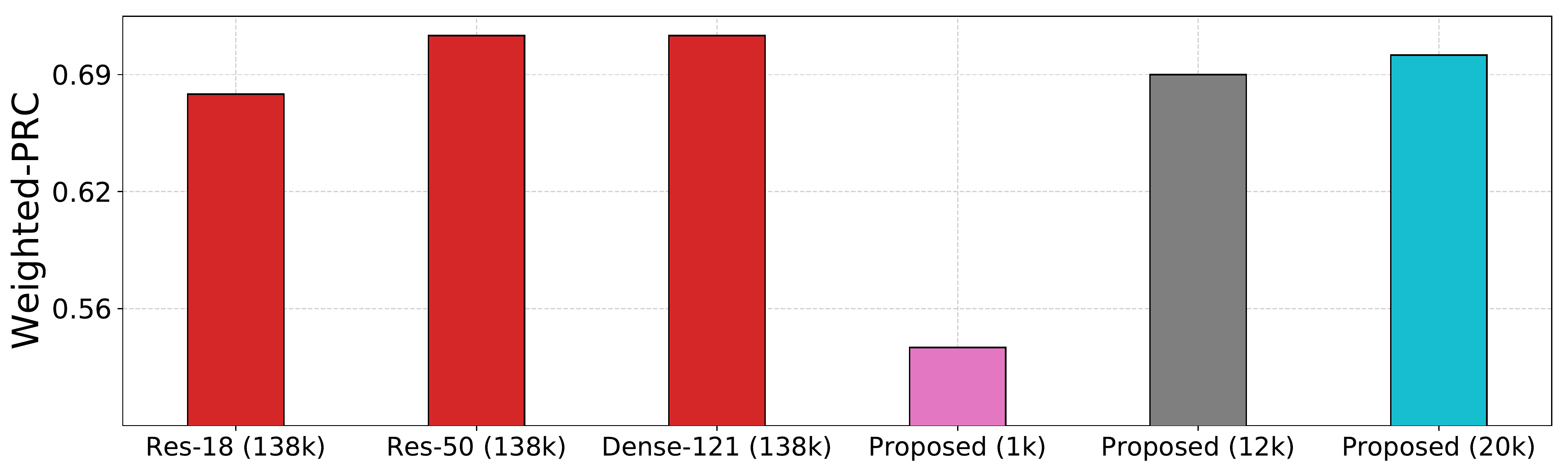}}
	\vspace{0.1in}
	\caption{Our ResNet-18 model trained with $85\%$ lesser labeled data matches the over-parameterized models trained on the full $138$k data.}
	\label{fig:comparison}
\end{figure*}

% \begin{figure*}[t]
% 	\centering
% 	\subfloat[$N_{\ell} = 1000$ and $N_{u} = 5000$]{\includegraphics[width=0.8\linewidth]{../figures/class_specific_auc_1000.pdf}} \\
% 	\subfloat[$N_{\ell} = 12500$ and $N_{u} = 15000$]{\includegraphics[width=0.8\linewidth]{../figures/class_specific_auc_12500.pdf}} 
% % 	\\
% % 	\subfloat[$N_{\ell} = 20000$ and $N_{u} = 30000$]{\includegraphics[width=0.80\linewidth]{../figures/class_specific_auc_20000.pdf}}
% 	\caption{Class-specific AUC achieved using different approaches for two few-shot scenarios. W-Aug.: weak augmentation, CT: confidence tempering, ST(L): self-training with labeled data, ST(U): self-training with additional unlabeled data.}
% 	\label{fig:teacher}
% \end{figure*}

\subsection{Key Findings}
\noindent \textbf{\textit{Mixup and confidence tempering provide significant gains.}} We find that the CT regularization, when combined with mixup and weak augmentation, produces crucial performance gains. For example, when $\mathrm{N_\ell}$=$12500$, W-AUC increases from $0.819$ to $0.833$ and W-PRC from $0.64$ to $0.673$. CT also improves the AUC for classes with low support while not compromising on those with high support. From the saliency maps for detecting different conditions in Figure \ref{fig:gradcam}, we find that the CT regularization leads to non-trivial probabilities even for negative findings, however, the evidences are from irrelevant parts of the image (e.g. organ boundary) thereby avoiding spurious correlations.

\noindent \textbf{\textit{Self-training with limited data matches full-shot training}.} Finally, incorporating the self-training strategy with additional unlabeled data boosts the performance even further. Surprisingly, using less than $10\%$ of the labeled data and $15$k unlabeled samples, our approach obtains a W-PRC score of $0.684$, outperforming  a state-of-the-art ResNet-18 trained on the full $138k$ set which scores $0.675$ (shown by the gray bar versus the first red bar in Fig.~\ref{fig:comparison}). More interestingly, as the number of labeled examples increases, the best performing ResNet-18 model obtained at $N_{\ell}=20000$ ($\sim 15\%$ of the total labeled data) is comparable to the significantly more complex ResNet-50 and DenseNet-121 models trained on the full data.

%% file: sections/discussion.tex
As deep learning methods continue to remain intensely hungry for data and compute infrastructure, it's real-world impact hinges on the availability of large-scale resources, thereby limiting democratization. Further, even when resources are available, the trained models can lack robustness and become vulnerable to shifts in data/task distributions and systemic biases. This phenomenon is especially prevalent in clinical domains, an area which is often challenged by limited data availability, cost of expert annotations, complexity of tasks, and privacy requirements. Hence, improving model generalization is imperative for effective adoption of AI tools in practice.

\noindent \textbf{Scientific Relevance.} In an effort towards democratizing AI for Computer-Aided Diagnosis (CAD) applications, we present a first of its kind deep learning framework to improve performance of classification models with limited labeled and unlabeled medical imaging data. Our proposed approach helps lay newer foundations for further exploration of learning methodologies to enhance generalization capabilities of clinical models. In particular, our finding that our approach required a shallower network and $85\%$ lesser data to match the performance of state-of-the-art neural network architectures is very important.

\noindent \textbf{Clinical Relevance.} We conduct a rigorous empirical study of our framework in classifying chest X-rays to detect $5$ prominent lung diseases. Given the extensive use of CXR data in early diagnostic settings, and the community-wide endeavor to curate, publish and benchmark large-scale public datasets through open-source competitions makes CXR classification a compelling problem to be solved. For example, the importance of building reliable CAD tools for CXR-based diagnosis has been strongly emphasized by the ongoing COVID-19 testing efforts. We believe that, by effectively leveraging small datasets to develop accurate predictive models, our approach provides new opportunities for rapid model development and deployment in clinical settings. 

\noindent \textbf{Future Work.} While the proposed work enables effective modeling with limited data, there are other inherent challenges in CXR classification, addressing which can lead to higher fidelity predictive models. The foremost challenge is the discrepancy in the sampling of different classes across datasets. For example, when the number of images with Pleural Effusion are significantly more than those with Atelectasis in the train set, and this does not hold true in an unseen test set, the model can perform poorly at detecting Atelectasis. This is commonly referred to as label distribution shifts, and our future work involves extending our framework to address this challenge. 
% Further, in clinical data, such shifts tend to cause a corresponding shift in the task being solved due to complex comorbidities. We foresee the need for new formulations that jointly tackle domain and task shifts in multi-label problems while being robust to outliers for achieving better generalization performances.